**Seth Golembeski[1]**
School of Aerospace Engineering,
Georgia Institute of Technology,
Atlanta, GA 30332, USA
sgolembeski3@gatech.edu

**Keith L. Gibson**
School of Aerospace Engineering,
Georgia Institute of Technology,
Atlanta, GA 30332, USA
kgibson43@gatech.edu

**Alexander Gross**
School of Aerospace Engineering,
Georgia Institute of Technology,
Atlanta, GA 30332, USA
Also with Sandia National Laboratories
agross36@gatech.edu

**Shreyas Kousik**
School of Mechanical Engineering,
Georgia Institute of Technology,
Atlanta, GA 30332, USA
shreyas.kousik@me.gatech.edu

**Anirban Mazumdar**
School of Mechanical Engineering,
Georgia Institute of Technology,
Atlanta, GA 30332, USA
Also with Sandia National Laboratories
anirban.mazumdar@me.gatech.edu

# Distributed Model-Based Diffusion: Finite Horizon Contraction under Bounded Delay

*Simultaneously optimizing the trajectories of multiple agents is a challenging problem plagued by nonlinearity, nonconvexity, and the curse of dimensionality. A collection of interacting aerial vehicles or self-driving cars in an intersection are examples of complex multi-agent systems that remain difficult to solve without many simplifying assumptions. The presence of communication latency between agents further increases the difficulty. In this paper, we analyze Distributed Model-Based Diffusion: a sampling-based Model-Predictive Control method suitable for highly nonlinear, nonconvex, nonsmooth, multi-agent systems. We prove contraction and robustness to latency for multi-agent, nonconvex problems, showing applicability to real-world constraints. We test the algorithm on a circleswap task, a cooperative medium-fidelity driving task, and in an aerial combat scenario. Despite the addition of latency, our algorithm improves circleswap makespan by 31% and increases aerial combat win rate by 25% compared to centralized Model-Based Diffusion.*

## 1 Introduction

Multi-robot systems offer great potential for a range of missions by providing robustness, complementary capabilities, and greater sensor coverage. However, coordinating multiple robots using Trajectory Optimization and Model-Predictive Control (MPC) remains challenging. While many algorithms exist to solve this class of problem, they often require strong assumptions about linearity, convexity, or differentiability [1,2].

Recently, Distributed Model-Based Diffusion (DMBD) [3] has been developed, which extends the powerful Model-Based Diffusion (MBD) nonconvex optimizer to distributed, multi-agent systems. We developed this work contemporaneously with and independently of [3], which did not prove convergence of the algorithm. In addition to such proof, our work builds upon this recent result by examining the critical challenge of degraded communication due to latency.

We show that DMBD globally contracts to a noise floor in finite time and provide robustness analysis that supports its utility under realistic network degradation. In particular, we make the following contributions:

- Proof of global finite-horizon contraction of DMBD for nonconvex objectives.
- Analysis of the above in the presence of latency between agents, ensuring that DMBD is still convergent when deployed on more realistic networked system models.
- Simulation results confirming that our convergence analysis holds in practice on systems of escalating complexity under degraded communication: 1) classic unicycle circleswap 2) multi-agent car racing and 3) multi-agent agile aerial combat games (dogfighting).

## 2 Background and Related Work

Multiagent problems deal with systems of $M$ agents where each agent has a disjoint partition $y_m \in \mathbb{R}^{d_m}$ of a decision variable $y \in \mathbb{R}^d$. Agents seek to solve $\min_y J(y)$ by solving their own component of the collective optimization problem. $J$ includes both direct costs and soft constraint penalties.

Classical trajectory optimization techniques typically require differentiability [1] and convexity or a valid initial guess [2] to guarantee convergence. Sampling-Based Model-Predictive Control (SBMPC) like Model-Predictive Path Integral (MPPI) and MBD [4] were developed to mitigate these problems. SBMPC avoids the convexity problem by using a principled Gaussian smoothing to progressively convexify problems into a homotopy of convex problems [5].

[1] Corresponding Author.

**2.1 Single-Agent Model-Based Diffusion.** Model-Based Diffusion (MBD) operates by directly sampling $J(Y)$ to make an estimate of $\nabla \log p_0(Y)$, which is a component of a reverse Stochastic Differential Equation (SDE) describing a noisy trajectory. Intuitively, this is gradient ascent on a smoothed surface, where the diffusion process has made the noised space convex. This convexification enables the global convergence in Sec. 3.

MBD is an iterative process with a total of $N$ steps. Since it is a reverse SDE, the optimization index $\cdot^{(i)}$ iterates from step $N$ to 1. For small temperature ($\lambda$) and infinite samples, the mode of this distribution is the optima of $J$ [4]; however, it is intractable to directly evaluate this value, hence we estimate it via the sampling process below.

The parameters $\alpha^{(i)}$ and $\sigma^{(i)}$, defined below, describe noise of the diffusion process. We use a geometric schedule for $\sigma^{(i)}$.

$$\sigma^{(i)} := \sqrt{\frac{1-\bar{\alpha}_i}{\bar{\alpha}^{(i)}}}, \quad \bar{\alpha}^{(i)} := \left(\prod_{k=1}^{i} \alpha_k\right) \forall i \in [1..N] \tag{1}$$

This is used to find the diffusion proposal distribution

$$\varphi^{(i)} := \mathcal{N}\left(x^{(i)}, (\sigma^{(i)})^2 I\right), \quad x^{(i)} := Y^{(i)}/\sqrt{\bar{\alpha}^{(i)}} \tag{2}$$

where $I$ is an identity matrix and $Y^{(i)}$ is the current iterate. The weighted mean $\bar{Y}^{(0)}$ of samples $\mathcal{Y}^{(i)} = \left\{y_1, \ldots, y_{n_s}\right\} \sim \varphi^{(i)}$ is then calculated from exponential weights, yielding the gradient estimator:

$$\bar{Y}^{(0)}(\mathcal{Y}^{(i)}) = \frac{\sum_{y\in\mathcal{Y}^{(i)}} y p_0(y)}{\sum_{y\in\mathcal{Y}^{(i)}} p_0(y)}, \quad p_0(y) \propto e^{-\frac{J(y)}{\lambda}}$$

$$\nabla_{Y^{(i)}} \log p_i(Y^{(i)}) \approx \frac{\sqrt{\bar{\alpha}^{(i)}}}{1-\bar{\alpha}^{(i)}} \bar{Y}^{(0)}(\mathcal{Y}^{(i)}) - \frac{Y^{(i)}}{1-\bar{\alpha}^{(i)}}. \tag{3}$$

Optimization then proceeds using gradient ascent:

$$x^{(i-1)} = x^{(i)} + \frac{\eta^{(i)}}{\sqrt{\bar{\alpha}^{(i)}}} \nabla_{Y^{(i)}} \log p_i\left(Y^{(i)}\right) \tag{4}$$

with step size $\eta^{(i)}$. Eqs. (3) and (4) form one entire optimization step. The original implementation of MBD uses $\eta^{(i)} = 1 - \bar{\alpha}^{(i)}$. Further derivation is available in [4,5].

**2.2 Related Methods.** MBD has been previously demonstrated for multi-agent systems [6]. MDOC [7] describes a multi-agent formulation using control barrier functions per agent. These algorithms operate on the joint space $y$ and not per-agent $y_m$. Other similar methods have previously been used for multi-agent planning, including Model Predictive Path Integral with agent prioritization [8], optimal reciprocal collision constraints [9], and MPPI aided by alternating direction method of multipliers (ADMM) [10]. Non-MPPI/MBD sampling based methods also exist, such as distributed subgradient methods [11], and zero-order stochastic methods [12].

**2.3 Distributed Model-Based Diffusion.** Distributed Model-Based Diffusion (DMBD) enables distributed, multi-agent optimization. Breaking the diffusion process into per-agent diffusion processes mitigates the sampling issues associated with increasing dimensionality, and, in physical systems, allows compute to be split amongst agents, improving cost and resilience to agent loss or network degradation. We first present the optimization algorithm (Alg. 1) and prove that it converges (Sec. 3). Finally, we discuss theoretical limitations (Sec. 3.4), providing the first proof of convergence for this distributed algorithm.

The DMBD algorithm extends standard MBD to $M$ agents. Each agent $m$ solves its own diffusion process optimizing $\min_{y_m} J(y)$ and periodically sharing information. The true diffusion state is denoted as $x^{(i)} = (x_{1,1}^{(i)}, x_{2,2}^{(i)}, \ldots x_{M,M}^{(i)})$. Each agent $m$ has an estimated state $\hat{x}_m^{(i)} = (\hat{x}_{m,1}^{(i)}, \hat{x}_{m,2}^{(i)} \ldots \hat{x}_{m,M}^{(i)})$ where subscripts indicate the agent that is making this estimate ($m$) and the agent it is estimating, respectively. Each agent draws samples of its own actions, evaluating the global cost function holding peers' actions constant.

This algorithm is run in parallel, with one instance per agent. Optionally, it can be used in a receding horizon sense. We augment the algorithm of [3] with timestamped messages to avoid duplicate message receipt or out-of-order messages due to latency. In the later experimental section (Sec. 4), we use a maximum latency of 50ms. This is intentionally conservative.

# 3 Convergence Analysis

In this section, we provide our main result: DMBD converges. To proceed, we first define the system and induced properties (Sec. 3.1.1), show that our method contracts stepwise (Sec. 3.2), and proceed to bound the error dynamics, with our key result in Theorem 3.6. Using these steps, we create a probabilistic likelihood of staying in the correct convex region, as well as a bound on RMS error.

**3.1 Assumptions, Definitions, and Setup for Proof.** This section describes the necessary preliminaries for the proof. We first discuss the multi-agent system, then the diffusion-based optimization.

**Algorithm 1** Distributed MBD

```
1: procedure AGENT j
2:     Input: x_j^(N) ~ N(0, σ_N^2)
3:     for i = N to 1 do
4:         for message g in async message queue from k do
5:             if time of g > time of last k update then
6:                 x̂_{j,k}^(i) ← g                                        ▷ Receive updates
7:             end if
8:         end for
9:         Y_{j,j}^(i) ~ N(x_{j,j}^(i), (σ^(i))^2)                      ▷ Collect samples
10:        Y^(i) ← [x̂_{j,1}^(i), ..., Y_{j,j}^(i), ..., x̂_{j,M}^(i)]
11:        x_{j,j}^(i-1) ← Equation (4) on Y^(i)
12:        Transmit new state x_{j,j}^(i-1) to all other agents
13:    end for
14: end procedure
```

*3.1.1 Distributed Multi-Agent System Setup.* We define the notation $x_m$ as agent $m$'s block, and $x_{-m}$ as all of the other blocks. Then, $J(x_m, x_{-m}) = J(x)$. Define $x^\star$ as the joint minimizer of $J$. The inner product is denoted as $\langle \cdot, \cdot \rangle$.

We require objective functions of the class described in [5, Asm. 4.1-4.2], which we restate in Asm. 1), i.e., functions with a smooth, convex core about their optima and subgaussian tails. MBD operates by expanding the convex region of the problem and then keeping iterates inside in a homotopy sense. We use this same property for our proof. A quadratic-coercive $J$ is sufficient to meet the subgaussian requirement and is standard as an input cost in many optimal control problems. When the solution lies away from discontinuity, the local smoothness and the standard second order optimality condition immediately imply the local convexity condition.

We assume the system has distributed cost structure with weak coupling and bounded latency:

**Assumption 1** (Distributed cost structure)**.** *There is a shared objective $J(x)$ with a unique minimizer $x^\star$. $J$ has a convex, smooth core and the induced $p_0$ is subgaussian in the tails. These assumptions are based on [5].*

**Assumption 2** (Weak coupling)**.** *The cost function $J$ has weak coupling between agents. Define the best response:*

$$B_m(x_{-m}) \coloneqq \arg\min_z J(z, x_{-m}). \tag{5}$$

*Then assume a coupling measurement $C \in \mathbb{R}^{M,M}$ such that*

$$\|B_m(v_{-m}) - x_m^\star\| \leq \sum_{j \neq m} C_{mj} \|v_j - x_j^\star\| \; \forall\, v \in \mathbb{R}^d \tag{6}$$

Put plainly, a best response's deviation from optima is governed by the magnitude of the deviation of other agents.

Next, we define the smoothed objective $G$ that is induced by the diffusion process.

**Definition 1** (Diffusion Objective)**.** *Define the blockwise smoothed objective, wherein $*$ is convolution and peer actions are held constant:*

$$G_m^{(i)}(x_m) \coloneqq -\lambda \log e^{-J(x_m, x_{-m})/\lambda} * \mathcal{N}(\cdot, (\sigma^{(i)})^2). \tag{7}$$

Next, since the agents do not have perfect communication, each must maintain an estimate of each others' states.

**Assumption 3** (Latency bound)**.** *Transmission delay is bounded: estimates $\hat{x}$ are no more than $\tau_{max}$ steps old.*

**Definition 2** (Distributed State Estimate Update)**.** *The multiagent is composed of $M$ agents with index $m \in [1..M]$. Represent the disjoint diffused state at step $i$ known by agent $m$ as $\hat{x}_m^{(i)} = (\hat{x}_{m,1}^{(i)}, \hat{x}_{m,2}^{(i)}, ..., \hat{x}_{m,M}^{(i)})$, wherein $\hat{x}_{m,j}^{(i)} \in \mathbb{R}^{d_m}$ is the $j$th agent's state estimated by agent $m$ at step $i$. Define each agents' dimension $d_m$ and total $d \coloneqq \sum d_m$.*

*The distributed update, with sampled gradient $\widehat{\nabla G}_m^{(i)}(x)$, is:*

$$x_{m,m}^{(i-1)} = x_{m,m}^{(i)} - \eta^{(i)} \widehat{\nabla G}_m^{(i)}(x) \tag{8}$$

$$-\widehat{\nabla G}_m^{(i)}(x) \coloneqq \frac{\bar{Y}^{(0)}(\mathcal{Y}^{(i)}) - x^{(i)}}{1 - \bar{\alpha}^{(i)}}. \tag{9}$$

Finally, the diffusion process induces an expanding local convexity on $G$, the radius of which expands linearly in $\sigma$.

**Lemma 3.1.** *$G_m^{(i)}$ is $a_m^{(i)}$-convex and $\beta_m^{(i)}$-smooth on a region $\mathcal{R}_m^{(i)}$ with radius $R_m^{(i)}$ centered on the minimizer.*

*Proof.* These properties follow from Asm. 1 and Def. 2 which satisfy [5, Thm 4.3], holding constant other agents' actions and sampling only own actions. □

*3.1.2 Diffusion Properties.* Since the convex ball expands (Lemma 3.1), we can trivially state that the most-expanded ball contains the initial iterate.

**Assumption 4** (Initial condition)**.** *The initial convex ball must contain the start point:* $x_m^{(N)} \in \mathcal{R}_m^{(N)}$.

Next, we impose bounds on step sizes.

**Assumption 5** (Step sizes)**.** *We require* $\eta$ *in the range*

$$\frac{1}{a_m^{(i)}}\left(1-\left(\frac{R_m^{(i-1)}}{R_m^{(i)}}\right)^2\right) < \eta^{(i)} < \frac{1}{\beta_m^{(i)}} \tag{10}$$

$$q_m^{(i)} := \sqrt{1-\eta^{(i)}a_m^{(i)}} \tag{11}$$

*When a geometric schedule* $(\sigma^{(i-1)})^2 = (\sigma^{(i)})^2\gamma$ *is used, the lower bound becomes* $\frac{1-\gamma}{a_m^{(i)}}$.

The lower bound is required to make Lemma 3.5 below well-defined, while the upper bound is standard for descent arguments; both are verifiable. Simply put, the step size must be large enough to stay in the shrinking convex region yet small enough to benefit from smoothness. The range must also be non-empty.

**3.2 Proving Stepwise Contraction.** This section proves stepwise contractive properties, starting with an optimality gap and then accounting for the various sources of error.

**Lemma 3.2** (Blockwise optimality gap)**.** *Define the smoothed best response:*

$$b_m^{(i)}(v_{-m}) := \arg\min_{z\in\mathcal{R}_m^{(i)}} G^{(i)}(z_m, v_{-m}). \tag{12}$$

*The smoothed best response gap bound is then*

$$\|b_m^{(i)}(v_{-m}) - B_m(v_{-m})\| \leq \psi_m^{(i)}, \tag{13}$$

*where* $\psi_m^{(i)}$ *is a function of the objective's condition number and other constant parameters.*

*Proof.* This result comes from [5, Thm. 4.4 (Optimality gap)] when applied with fixed peer actions. For space, we omit the exact calculable definition of $\psi$. □

**Lemma 3.3** (Stepwise contraction)**.** *At each step, the error from the smoothed best response (Eq. (12)) contracts up to gradient estimator error:*

$$\|x_m^{(i-1)} - b_m^{(i)}(\hat{x}_{m,-m}^{(i)})\| \leq q_m^{(i)}\|x_m^{(i)} - b_m^{(i)}(\hat{x}_{m,-m}^{(i)})\| + \eta^{(i)}\|\zeta_m^{(i)}\|$$

*Proof.* Suppose $x^{(i)} \in \mathcal{R}^{(i)}$. Denote the gradient estimator error $\zeta_m^{(i)} = \widehat{\nabla G}_m^{(i)} - \nabla G_m^{(i)}$. Then define the deterministic update,

$$\bar{x}_m^{(i-1)} = x_m^{(i)} - \eta^{(i)}(\nabla G_m^{(i)}). \tag{14}$$

A standard gradient contraction (e.g., [13, Thm. 3.6]) states:

$$\|\bar{x}_m^{(i-1)} - b_m^{(i)}(\hat{x}_{m,-m}^{(i)})\| \leq q_m^{(i)}\|x_m^{(i)} - b_m^{(i)}(\hat{x}_{m,-m}^{(i)})\| \tag{15}$$

Then using $x_m^{(i-1)} = \bar{x}_m^{(i-1)} - \eta^{(i)}\zeta_m^{(i)}$ yields the proof by triangle inequality. □

**3.3 Bounding the Error Dynamics.** We now describe the RMS error dynamics of the diffusion system during optimization, culminating in a finite-horizon bound in Theorem 3.6.

**Lemma 3.4** (RMS Recursive Bound)**.** *Denote the error*

$$r_m^{(i)} := \left(\mathbb{E}\left[\|x_m^{(i)} - x_m^\star\|^2\right]\right)^{1/2}. \tag{16}$$

*Then* $r$ *is recursively upper-bounded:*

$$r^{(i-1)} \leq A^{(i)}r^{(i)} + D^{(i)} \textit{and} \tag{17}$$

$$\begin{aligned} r_m^{(i-1)} \leq{}& q_m^{(i)}r_m^{(i)} + \eta^{(i)}\sqrt{v_m^{(i)}} + (1+q_m^{(i)})\big(\psi_m^{(i)} + \\ &+ \sum C_{mj}r_j^{(i)} + \sum C_{mj}l_{m,j}^{(i)}\big), \end{aligned} \tag{18}$$

*where $A^{(i)} \in \mathbb{R}^{M,M}$ is the state transition matrix, and $D^{(i)} \in \mathbb{R}^{M}$ is the suboptimality forcing bound, given by*

$$A^{(i)}_{m,m} = q^{(i)}_m, A^{(i)}_{m,j} = (1 + q^{(i)}_m)C_{m,j}, j \neq m \tag{19}$$

$$D^{(i)}_m = (1 + q^{(i)}_m)\left(\psi^{(i)}_m + \sum C_{m,j} l^{(i)}_{m,j}\right) + \eta^{(i)}\sqrt{\nu^{(i)}_m} \tag{20}$$

$$\nu^{(i)}_m = \frac{\lambda^2 d_m}{n_s}\left(\frac{V_{-1}}{(\sigma^{(i)})^2} + V_0\left(\frac{L^{(i)}_m}{\lambda}\right)^2 + V_1\left(\frac{L^{(i)}_m}{\lambda}\right)^4 \sigma^2\right) \tag{21}$$

$$l^{(i)}_{m,j} = \sum_{k=i+1}^{\min(N,i+\tau_{max})} \eta^{(k)}\sqrt{2(\beta^{(k)}_j R^{(k)}_j)^2 + 2\nu^{(k)}_j}. \tag{22}$$

*Proof.* Yi [5, C.1] gives us

$$\mathbb{E}\,\|\widehat{\nabla G}^{(i)}_m(x) - \nabla G^{(i)}_m(x)\|^2_2 \leq \nu^{(i)}_m, \tag{23}$$

where each $\nu^{(i)}_m$ is constant and $G^{(i)}_m$ is $L^{(i)}_m$-Lipschitz. The locally smooth region bounds the gradient by:

$$\mathbb{E}\,\|\nabla G^{(i)}_m(x) + \zeta^{(i)}_m\|^2 \leq 2(\beta^{(i)}_m R^{(i)}_m)^2 + 2\nu^{(i)}_m. \tag{24}$$

We then use Minkowski's inequality to yield the estimation error bound $l$:

$$\left(\mathbb{E}\left[\|\hat{x}^{(i)}_{m,j} - x^{(i)}_{j,j}\|^2\right]\right)^{1/2} \leq l^{(i)}_{m,j}. \tag{25}$$

Write the best-response error and bound with Lemma 3.2:

$$\begin{aligned} b^{(i)}_m(\hat{x}^{(i)}_{m,-m}) - x^\star_m = b^{(i)}_m(\hat{x}^{(i)}_{m,-m}) - B_m(\hat{x}^{(i)}_{m,-m}) + \\ B_m(\hat{x}^{(i)}_{m,-m}) - B_m(x^\star_{-m}) \end{aligned} \tag{26}$$

$$\begin{aligned} \|b^{(i)}_m(\hat{x}^{(i)}_{m,-m}) - x^\star_m\| \leq \psi^{(i)}_m + \sum C_{mj}\|x^{(i)}_j - x^\star_j\| + \\ \sum C_{mj}\|\hat{x}^{(i)}_{m,j} - x^{(i)}_j\|. \end{aligned} \tag{27}$$

Write the next iterate error, apply the triangle inequality, and bound with Lemma 3.3:

$$\begin{aligned} \|x^{(i-1)}_m - x^\star_m\| &= \|x^{(i-1)}_m - b^{(i)}_m(\hat{x}^{(i)}_{m,-m}) + \\ &+ b^{(i)}_m(\hat{x}^{(i)}_{m,-m}) - x^\star_m\| \\ \|x^{(i-1)}_m - x^\star_m\| &\leq q^{(i)}_m\|x^{(i)}_m - b^{(i)}_m(\hat{x}^{(i)}_{m,-m})\| + \eta^{(i)}\|\zeta^{(i)}_m\| \\ &+ \|b^{(i)}_m(\hat{x}^{(i)}_{m,-m}) - x^\star_m\| \end{aligned} \tag{28}$$

By a similar triangle inequality,

$$\|x^{(i)}_m - b^{(i)}_m(\hat{x}^{(i)}_{m,-m})\| \leq \|x^{(i)}_m - x^\star_m\| + \|b^{(i)}_m(\hat{x}^{(i)}_{m,-m}) - x^\star_m\| \tag{29}$$

$$\begin{aligned} \|x^{(i-1)}_m - x^\star_m\| \leq &(1 + q^{(i)}_m)\|b^{(i)}_m(\hat{x}^{(i)}_{m,-m}) - x^\star_m\| + \\ &\eta^{(i)}\|\zeta^{(i)}_m\| + q^{(i)}_m\|x^{(i)}_m - x^\star_m\| \end{aligned} \tag{30}$$

Apply Minkowski's inequality and use Eq. (27) with the dynamical system definition to obtain the RMS error. □

The previous lemma describes the influence of the various sources of error on the system and how they interact with the contractive tendency. This is necessary to fully quantify the performance of the optimizer.

Next, we prove that the iterates remain in the convex ball. Diffusion optimality hinges on the convexification process; without it, we cannot easily describe what would happen to iterates outside the ball.

**Definition 3** (Iterate in-bounds event)**.** *Define the in-basin event indicating when iterates are in the convex ball.*

$$\epsilon^{(i)} := \bigcap_{k=i}^{N}\left\{x^{(k)}_m \in \mathcal{R}^{(k)}_m \forall m\right\}. \tag{31}$$

**Lemma 3.5** (Boundedness of DMBD)**.** *The diffusion process remains in the convex region* $\mathcal{R}_m^{(i)}$ *with probability* $p_b$*:*

$$p_b \geq \prod_{k=1}^{N} \left(1 - \sum_{m=1}^{M} \min(p_{e,m}^{(k)}, 1)\right), \quad \textit{where} \tag{32}$$

$$p_{e,m}^{(i)} \leq \frac{\left(\eta^{(i)}\sqrt{\nu_m^{(i)}}+\delta_m^{(i)}\right)^2}{\left(R_m^{(i-1)}-q_m^{(i)}R_m^{(i)}\right)^2}. \tag{33}$$

*Proof.* Define the stepwise drift

$$\Delta_m^{(i)} := \|b_m^{(i)}(\hat{x}_{m,-m}^{(i)}) - b_m^{(i-1)}(\hat{x}_{m-m}^{(i-1)})\| \tag{34}$$

$$\delta_m^{(i)} := (\mathbb{E}\left[(\Delta_m^{(i)})^2\right])^{\frac{1}{2}}. \tag{35}$$

In the worst case, $\delta$ can be bounded as

$$\delta_m^{(i)} \leq \sum_{k\in\{i,i-1\}} \psi_m^{(k)} + \sum_{j\neq m} C_{mj}(r_j^{(k)} + l_{m,j}^{(k)}) \tag{36}$$

This can be tightened for special cases, like a smooth $J$. Then,

$$\|x_m^{(i-1)} - b_m^{(i-1)}(x_m^{(i-1)})\| \leq \|x_m^{(i-1)} - b_m^{(i)}(x_m^{(i)})\| + \tag{37}$$

$$+ \|b_m^{(i)}(x_m^{(i)}) - b_m^{(i-1)}(x_m^{(i-1)})\|$$

$$\leq q_m^{(i)} R_m^{(i)} + \eta^{(i)}\|\zeta_m^{(i)}\| + \Delta_m^{(i)}. \tag{38}$$

Finally, the exit probability can be defined as:

$$p_{e,m}^{(i)} = \mathbb{P}\left[\eta^{(i)}\|\zeta_m^{(i)}\| + \Delta_m^{(i)} > R_m^{(i-1)} - q_m^{(i)} R_m^{(i)} | \epsilon^{(i)}\right]. \tag{39}$$

Using Markov's inequality squared completes the proof. □

Now, we can both bound the error dynamics and prove that iterates are in the convex region, which was a prerequisite for the error dynamics lemma. Combining them, we get our capstone result:

**Theorem 3.6** (Finite-Horizon RMS Error Bound for DMBD)**.** *By the above assumptions, the RMS error dynamics of DMBD are upper bounded by the recursion:*

$$\|r^0\|_\infty \leq \|r^{(N)}\|_\infty \prod_{i=1}^{N} \rho^{(i)} + \sum_{j=1}^{N} \left(\|D^j\|_\infty \prod_{k=1}^{j-1} \rho^{(k)}\right) \tag{40}$$

*with probability at least* $p_b$*, where* $\rho^{(i)} := \|A^{(i)}\|_\infty$.

*Proof.* Start by induction to show that iterates remain in the convex region. The base case $x_m^{(N)} \in \mathbb{R}_m^{(N)}$ is from Asm. 4. On $\epsilon^{(i)}$, the in-basin requirements of Lemmata 3.1 and 3.5 are met. Then, $\mathbb{P}(\epsilon^{(i-1)}|\epsilon^{(i)}) \geq 1 - \sum_m^M (p_{e,m}^{(i)})$. By induced-norm properties, we have the one-step contraction:

$$\begin{aligned} \|r^{(i-1)}\|_\infty &\leq \|A^{(i)}r^{(i)} + D^{(i)}\|_\infty \\ \|r^{(i-1)}\|_\infty &\leq \rho^{(i)}\|r^{(i)}\|_\infty + \|D^{(i)}\|_\infty. \end{aligned} \tag{41}$$

Applying the union bound over all steps and agents gives $\mathbb{P}(\epsilon^{(0)}) > p_b$. On this event, iterating over Eq. (41) yields Eq. (40). □

Using the above theorem, we can provide a rough upper bound on error and describe the relationship between contraction and suboptimality forcing.

**Corollary 3.6.1** (Contraction conditions)**.** *Suppose there exists a worst-case suboptimality forcing* $\bar{D}$ *such that* $\|D^{(i)}\|_\infty \leq \bar{D}$*. Suppose that*

$$\bar{\rho} := \max_{i,m} \left(q_m^{(i)} + (1 + q_m^{(i)}) \sum_{j\neq m} C_{m,j}\right) < 1, \tag{42}$$

*which can be directly verified. Then the error is bounded:*

$$\|r^{(0)}\|_\infty \leq \bar{\rho}^N \|r^{(N)}\|_\infty + \frac{1-\bar{\rho}^N}{1-\bar{\rho}}\bar{D}. \tag{43}$$

*Proof.* The proof is immediate from the product and geometric sum identity applied to Eq. (40). □

This corollary indicates that there is a tradeoff between contraction coefficient and the terminal noise floor.

**3.4 Theory limitations.** The above theory is limited to problems with a smooth, convex core, shared cost function, weak coupling, and bounded latency. The core smoothness and convexity can be verified numerically for a given $J$, while the weak coupling condition can be validated numerically or calculated directly from any twice-differentiable objective.

## 4 Experiment Design

We test DMBD in three environments of increasing difficulty to illustrate its benefits and assess its limits. Here we present the environments and experiment parameters, then Sec. 5 presents results and discussion.

First, we test a standard benchmark "Circleswap" environment from the multi-agent literature [14]. Next, we show that our method extends to more complex, competitive dynamics with "Multi-Car Racing" [15]. Finally, we evaluate our algorithm in "Simulated Aerial Combat", wherein a team of up to 6 agents (each a 6 degree-of-freedom rigidbody, $d_m = 1440$) must coordinate while competing against another team, controlled by a different algorithm [16].

**4.1 Common Experiment Design Features.** For all problems, define $n_r$ as the total number of samples available. In order to maintain computational fairness in the distributed case, each agent's samples are reduced to $n_s = n_r/M$, while the centralized optimizer maintains $n_s = n_r$. This maintains an exactly equal amount of rollout computation between cases, regardless of parallelization. All experiments are conducted with the parameters given in Table 1.

**Table 1 Optimizer parameters**

| | $\lambda$ | $N$ | $n_r$ | $\sigma^N$ | $\sigma^{(0)}$ | $\tau_{\max}$, steps |
|---|---|---|---|---|---|---|
| Circleswap | 0.1 | 150 | 300 | 0.6 | $10^{-2}$ | [0,10,20,30,50] |
| Racing, $\sigma+$ | 50 | 100 | 50 | 0.6 | $10^{-2}$ | [0,10,20,30,50] |
| Dogfighting | 0.1 | 50 | 100 | 0.6 | $10^{-2}$ | [0,10, 20, 30] |

We calculate the latency in simulations $\tau$ as a uniform distribution, $\tau \sim U(0, \tau_{\max})$, where $\tau_{\max}$ is swept in the ranges of the above table.

**4.2 Circleswap.** This problem uses a standard unicycle model with a fixed velocity; the control input is angular velocity.

This problem seeks to move $M$ agents from starting positions on the circumference of a circle ($r = 2$ meters) to goals diametrically opposite without collision. Agents are initialized with random headings $\theta \sim U(-\pi, \pi)$. We use a timestep of 0.3 seconds and an identical replanning rate, with a lookahead time of 40 timesteps in a receding horizon scheme. The dynamics are configured with $v = 0.2\frac{m}{s}, \omega_{\max} = 0.2\frac{\text{rad}}{s}$.

We use a cost function that penalizes squared error from the goal, assigns goal reward $w_g = 100$, and a large collision penalty $w_c = 100$. We run 100 trials per configuration.

**4.3 Multi-Car Racing.** As a second problem, we examine the multi-car racing problem from [15], which seeks to maximize the speed of $M$ medium-fidelity car models traveling around a race track while avoiding collision and maintaining a particular formation. We use a physics timestep of 10ms and a control timestep of 100ms with a lookahead horizon of 5s. The experiment is evaluated for a total of 25 rolled-out seconds. Cars are rewarded for velocity and maintaining a prescribed formation. They are penalized for deviating from the centerline, having large sideslip, or exiting the track. Further details and results for this environment can be found in [15,17]. We run 100 trials per configuration.

**4.4 Simulated Aerial Combat.** We evaluate the algorithm in a six-degree-of-freedom air combat simulation environment featuring F-16 aircraft. This environment is adapted from Gross [16]. Aircraft have a cone of fire allowing them to damage other agents. If an agent sustains sufficient damage, it is considered eliminated. If all hostile (friendly) agents are eliminated, it is considered a win (loss). Friendly agents receive reward from damaging hostile agents and are penalized for being damaged. A large reward is given for a win, and a small reward-shaping is applied for pointing near an adversary. Full details of the reward function are available in [16].

Blue agents use MBD, DMBD, or Pure Pursuit as guidance. They have complete knowledge of the adversary state and the adversary guidance. We accept this limitation in order to restrict the scope of experimentation to only our guidance law. Each team's initial heading is varied in 90° increments, giving a total of 16 starting conditions capturing head-on, tail chase, and other varieties of engagements. We run 160 trials per optimizer parameter configuration.

## 5 Results

We find experimentally that DMBD outperforms MBD and baselines in all three experiments. We also find that DMBD is robust to latency, as predicted from our mathematical bounds in Theorem 3.6. Specifically, DMBD performance decreases gradually with added latency, and DMBD generally outperforms centralized MBD even in the face of latency.

**5.1 Circleswap.** DMBD performs significantly better than centralized MBD, showing a significantly shallower slope of makespan with increasing agents. While the overall shape is similar, we see what is an approximately linear scaling.

Thanks to the relatively weak coupling in this problem, the latency has negligible impact on the makespan. All latency configurations have comparable makespan.

**5.2 Multi-Car Racing.** The multi-car racing environment shows similarly improved results. We present the results in terms of normalized reward, which is defined as $-J/(TM)$ in order to account for varying number of agents. We find that the best distributed configuration outperforms the centralized solver. DMBD shows a shallower degradation slope with increased agent count. This problem has the strongest coupling of the three used, and thus has the strongest degradation with latency. Nonetheless, DMBD degrades gracefully. With $\tau = 50$ – half of the optimization window – performance is still only mildly degraded from standard MBD (which is centralized and has no latency).

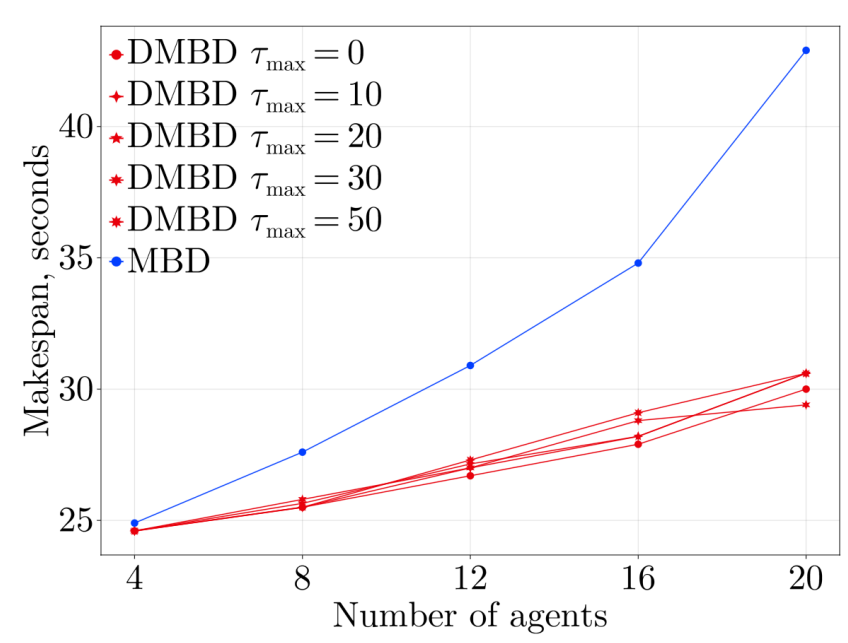

**Fig. 1 Circleswap makespan.**

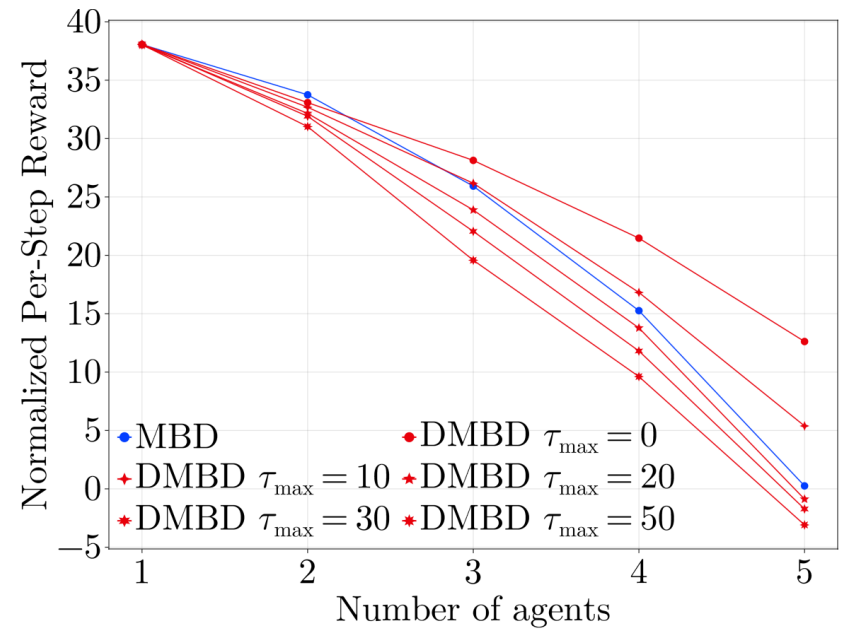

**Fig. 2 Multi-car racing normalized reward.**

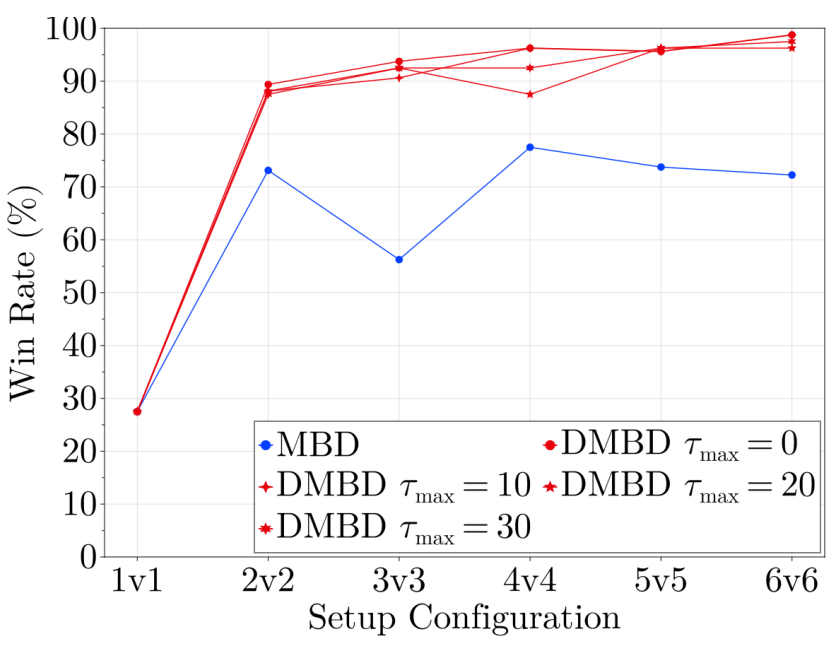

**Fig. 3 Dogfighting win rate.**

**5.3 Dogfighting.** The dogfighting environment similarly shows improvement with decentralization. We find an improvement in win rate of approximately 25% in the 6 v 6 engagement. Win rate in this problem is largely unaffected by latency.

## 6 Conclusion

Decentralization of a sampling-based Model-Predictive Control algorithm like MBD provides strong performance for multi-agent path planning. In all cases, we show that DMBD outperforms baseline MBD. These results provide additional validation to [3]

We also show mathematically that DMBD is stable despite increasing latency or network degradation. We demonstrate improvements in makespan, reward, and win rate across a diverse set of experiments as well as providing probabilistic bounds on performance. We also demonstrate strong robustness to simulated network latency in the above experiments. Experiments show that the latency degradation is most strongly influenced by the coupling strength, which aligns with our mathematical analysis.

## Acknowledgment

Sandia National Laboratories is a multimission laboratory managed and operated by National Technology & Engineering Solutions of Sandia, LLC, a wholly owned subsidiary of Honeywell International Inc., for the U.S. Department of Energy's National Nuclear Security Administration under contract DE-NA0003525, SAND2026-25752R. This paper describes objective technical results and analysis. Any subjective views or opinions that might be expressed in the paper do not necessarily represent the views of the U.S. Department of Energy or the United States Government. This work is sponsored by the Air Force Research Laboratory, Munitions Directorate (RWTA), Eglin AFB, FL under contract A2308-057-089-003242. Opinions, findings and conclusions, or recommendations are those of the authors and do not necessarily reflect the views of the sponsoring agencies. This material is based upon work supported by the National Science Foundation Graduate Research Fellowship Program under Grant No. DGE-2039655. Any opinions, findings, and conclusions or recommendations expressed in this material are those of the author(s) and do not necessarily reflect the views of the National Science Foundation.